\documentclass[11pt,a4paper]{article}
\usepackage[hyperref]{emnlp2020}
\usepackage{times}
\usepackage{latexsym}
\usepackage{makecell}
\usepackage{graphicx}
\usepackage{adjustbox}
\usepackage{float}
\usepackage{amssymb}
\usepackage{natbib}
\usepackage{epstopdf} 

\usepackage{amsmath,amsfonts}

\usepackage{microtype}

\aclfinalcopy 

\title{Not-NUTs at W-NUT 2020 Task 2: A BERT-based System in Identifying Informative COVID-19 English Tweets}

\author{Thai Hoang \\
  University of Washington\\
  \texttt{qthai912@cs.washington.edu} \\\And
  Phuong Vu $^*$ \\
  University of Rochester\\
  \texttt{pvu3@u.rochester.edu} \\}

\date{}

\begin{document}

\maketitle
\def\thefootnote{*}\footnotetext{Equal contribution with the first author}

\begin{abstract}

As of 2020 when the COVID-19 pandemic is full-blown on a global scale, people's need to have access to legitimate information regarding COVID-19 is more urgent than ever, especially via online media where the abundance of irrelevant information overshadows the more informative ones. In response to such, we proposed a model that, given an English tweet, automatically identifies whether that tweet bears informative content regarding COVID-19 or not. By ensembling different BERTweet model configurations, we have achieved competitive results that are only shy of those by top performing teams by roughly 1\% in terms of F1 score on the informative class. In the post-competition period, we have also experimented with various other approaches that potentially boost generalization to a new dataset. Our repository can be found in the following link:
\url{https://github.com/quocthai9120/W-NUT-2020-Shared-Task-2}

\end{abstract}

\section{Introduction}

Following the rise of smart technology and an increasingly wide coverage of Internet, social network websites are becoming ubiquitous these days. Besides serving as a platform for various types of entertainment, social media is particularly helpful in spreading information, and such can be leveraged to keep the majority of its users well-informed amidst a natural disaster or a pandemic like COVID-19. One major advantage of sourcing information via social media is that all information is updated in real-time. Any person with a social media account can post or share information instantly at the moment he/she witness a noteworthy event. This is a much faster way to obtain information compared to reading newspaper, watching the news on TV, or viewing other official source of information since most tend to be updated only at mid-day or at the end of day. Nevertheless, information on social media platforms is mostly not verified, heavily opinionated towards the person who posted it, and at worst, completely inaccurate. This highlights the need for a system that can automatically identify legitimate information from the huge pool of information. 

In order to address the aforementioned need for such a system, in this paper we attempt to tackle the WNUT 2020 Task 2: Identification of Informative COVID-19 English Tweets \cite{covid19tweet}. As stated in the task's description paper, this task requires its participants to build and refine systems that, given an English Tweet carrying COVID-19-related content, automatically classify whether it is informative or not. In the context of this shared task, being informative is defined as bearing information regarding suspected, confirmed, recovered or death cases related to COVID-19 as well as location or travel history of these cases. 

\section{Related work}

Text classification is a simple but practical task in the field of natural language processing. Early models such as Naive Bayes, Logistic Regression, and Support Vector Machine are widely known and used as a headstart for experimenting classification tasks due to their simplicity and fast training time while still able to achieve a reasonable performance.

The rise of modern neural network brings deep learning to the classification tasks within the language processing field as it helps induce features for learning. Further development of recurrent networks gives us the ability to deal with sequences of varied lengths, which improves the performance of text classification to a great extent.

While classifying texts, it is essential to make the machine understand deeply the characteristics of input sequences. Because of that, having a well-performing system that embed text sequences is an important prerequisite in building a good model for text classification.

Recently, pre-trained language models let us achieve high quality text embeddings, which then can be used for further downstream tasks. For language processing, the most famous pre-trained contextual language models recently are BERT \cite{1810.04805}, ELMOs \cite{1802.05365}, and XL-NET \cite{1906.08237}.

\section{System Description}

We use the pre-trained language model BERTweet \cite{BERTweet}, an English Tweet domain-specific model inspired by the original BERT model \cite{1810.04805}, as the core for our system (more details will be discussed later). To accomplish the task of identifying informativeness of COVID-19 English Tweets, we attach a classification block on top of our BERTweet block, which is a combination of one or more linear layers. Figure 1 indicates the high level detail of our system.

\begin{figure}[ht]
    \centering
    \includegraphics[width=\columnwidth]{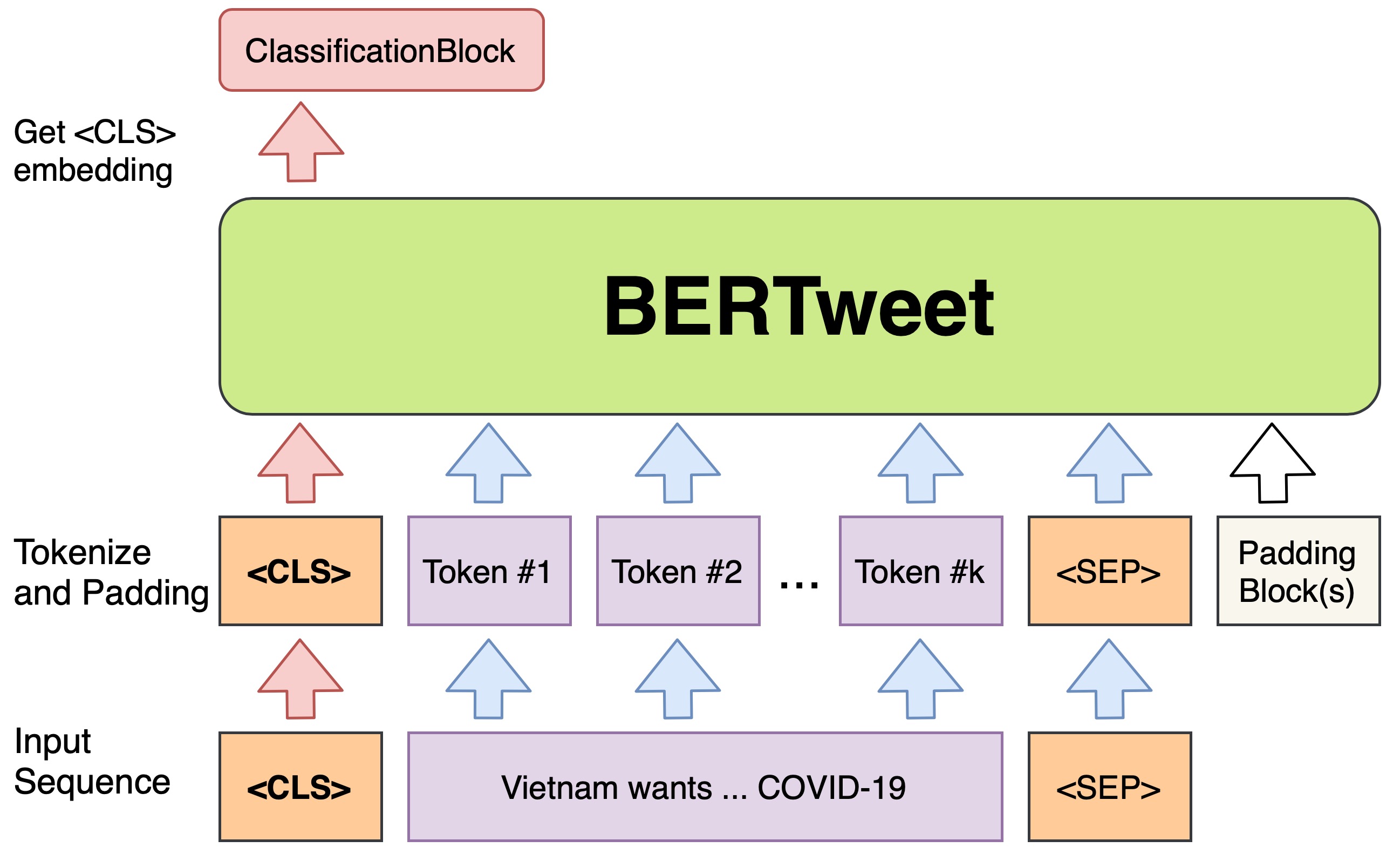}
    \caption{An overview of our model for identify Informative COVID-19 English Tweets}
    \label{fig:model}
\end{figure}

\subsection{BERTweet}
BERTweet \cite{BERTweet} is a large-scale language model pre-trained for English Tweets. Because of its nature of being a domain-specific model, BERTweet has achieved state-of-the-art performances on many downstream Tweet NLP tasks such as part-of-speech tagging, named entity recognition, and text classification, outperformed top models such as RoBERTa-base \cite{1907.11692} and XLM-R-base \cite{1911.02116}. Trained on 845M Tweets streamed from 01/2012 to 08/2019 and 5M Tweets related the COVID-19 pandemic as pre-training resources, BERTweet has an advantage compares to other models for classifying COVID-19 related English Tweets.

\subsubsection{Input Processing}
Before feeding into the BERTweet model, we first tokenize input sequences with BPE Tokenizer \cite{1508.07909}, then pad the input sequences with the $\texttt{[CLS]}$ and $\texttt{[SEP]}$ tokens at their beginning and ending positions. To ensure all sequences have uniform length, we also add padding blocks at the end of the input sequences. The tokenized and padded input sequences are then fed directly into the Transformer block to retrieve contextualized sequence embeddings.

\subsubsection{Embedding Extraction}
Each Transformer layer within BERTweet model learns different information. We experiment different ways of extracting the pooled token from our BERTweet model, which corresponds to the encoded $\texttt{[CLS]}$ token in our implementation, to analyze the performance on this downstream task. More detail would be discussed in the ``Experiments'' section.

\subsubsection{Global Local BERTweet}
By a close manual inspection of the dataset provided for the task, we realize that many Tweets have noteworthy information at some particular parts. Follow that reasoning, paying special attention to smaller parts of the Tweets is also important. Inspired by that idea, we propose a method to train 3 BERTweet models simultaneously: one for getting contextualized embeddings over the whole input sequences, one for getting embeddings over the first part of the Tweets, and one for getting embeddings over the remaining part. The pooled token from each model would then be extracted and concatenated together for the system to learn both global and local information of the Tweets. Please refer to Figure 2 for a visualization of the model.

\begin{figure}[ht]
    \centering
    \includegraphics[width=\columnwidth]{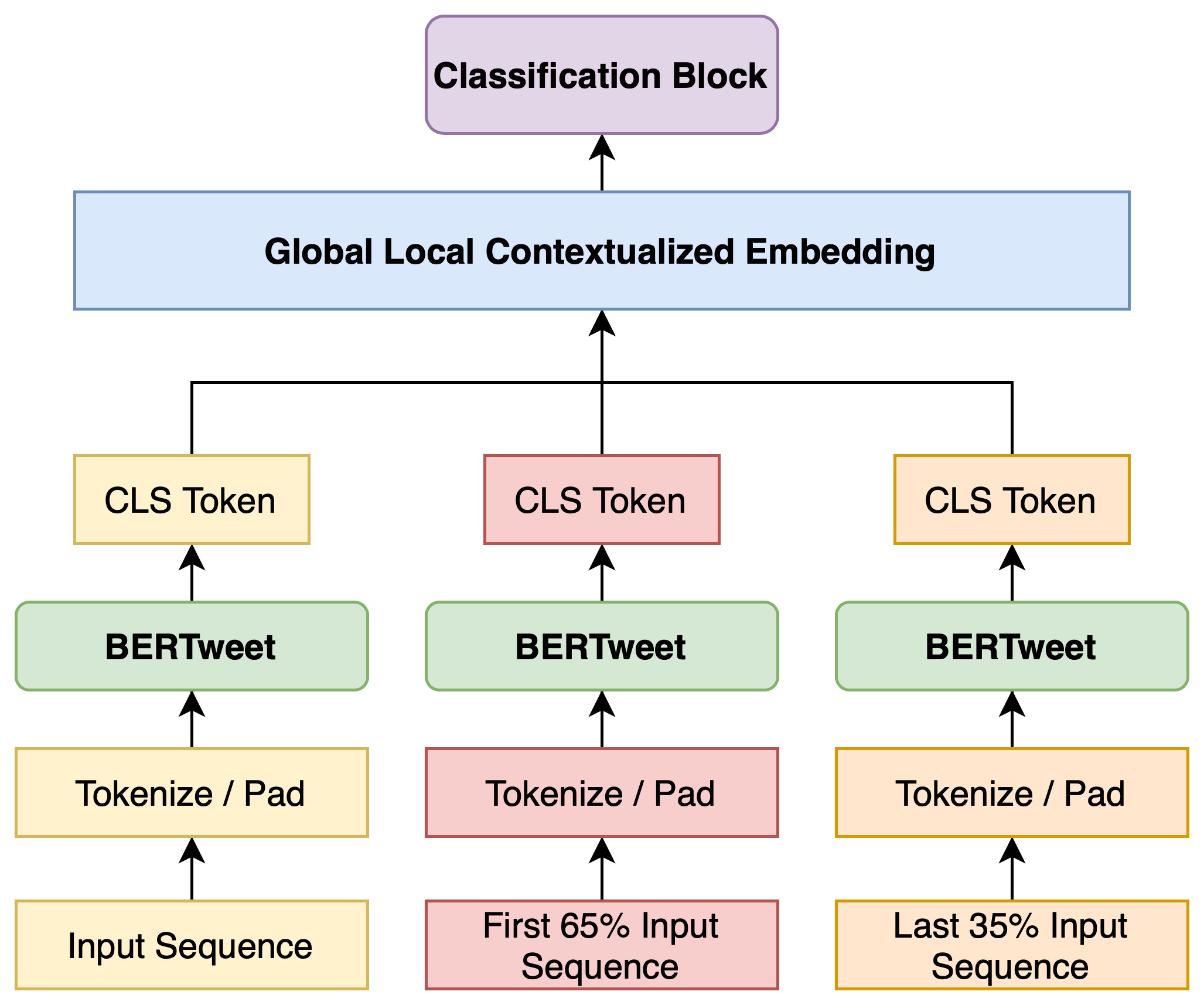}
    \caption{Global Local BERTweet Model}
    \label{fig:model}
\end{figure}

\subsection{Classification Block}
The classification block contains one or more linear layers stacked on top of each other. The final layer is then used to classify whether a Tweet is informative or not.

\section{Experiments}
\subsection{Dataset}
We use the dataset released by the competition organizer, consisting of 10,000 COVID-19 English Tweet. Each Tweet in the dataset is annotated by 3 annotators independently, and the overall inter-annotator agreement score of Fleiss' Kappa is 0.818. The dataset is then divided into 3 distinct set for training, validation, and testing, with the ratio of 70/10/20, respectively. Table 1 shows the division of the dataset.

\begin{table}[ht]
    \centering
    \begin{tabular}{|c|c|c|}
        \hline
         & \bf{Informative} & \bf{Uninformative} \\ \hline
        Training Set & 3303 & 3697 \\ \hline
        Validation Set & 472 & 528 \\ \hline
        Test Set & 944 & 1056 \\ \hline
    \end{tabular}
    \caption{Number of Tweets of each category in the original dataset}
    \label{tab:Dataset}
\end{table}

\subsubsection{Re-splitting Data}
During the final evaluation phrase, we re-split the dataset by combining training and validation sets then dividing randomly with the ratio of 90/10. The test set is not modified.


\subsection{Implementation}
\subsubsection{Main Library and Framework}
We mainly rely on the \texttt{transformers} library \cite{1910.03771} with \texttt{PyTorch} framework \cite{paszke2017automatic} to run our code.

\subsubsection{Two-Phrase Training}
We divide the training process into two phrases. In the first phrase, we freeze all the BERTweet paramaters to train the classification block. In the second phrase, we then unfreeze all parameters in our end-to-end model for finetuning.

\subsubsection{Optimizer}
For all models belonging to the scope of our project, we utilized the AdamW optimizer as implemented in the \texttt{transformers} library. This is a third-party implementation of the algorithm originally proposed in the paper named Decoupled Weight Decay Regularization \cite{loshchilov2019decoupled}

\subsubsection{Hyperparameters Configuration}
The max length for padding input sequences before feeding into the BERTweet model is set to be 256. We trained our models on 1 NVIDIA Tesla V100 and 1 NVIDIA GeForce RTX 2080 Ti using batch size of 16 and 32 alternatively. We use an initial learning rate of $5e-4$ in 12 epochs for the first phrase and $4e-5$ in 6 epochs for the second phrase of training along with linear learning rate decay then choose the best checkpoint.

\subsection{Model Performance}
\subsubsection{Baselines}
We pre-process input data by tokenizing the data, record the count of occurrences of each token in a matrix then transform such count matrix into a tf-idf representation. To do so, we use CountVectorizer() and TfidfTransformer() as implemented in \texttt{sklearn} \cite{scikit-learn}. We then use 3 different classifiers, namely SVM, Naive Bayes and Logistic Regression, to get results on the original validation set. We acknowledge that the performance of these baselines are relatively poor; nevertheless, it is a trade-off between accuracy and efficiency since follow a non-deep learning approach which does not require much time regarding training and finetuning. 

\begin{table}[ht]
    \centering
    \begin{tabular}{|c|c|}
        \hline
        \bf{Models} & \bf{F1 Score} \\\hline
        Logistic Regression & \textbf{0.7827} \\\hline
        Naive Bayes & 0.7486  \\\hline
        Support Vector Machine & 0.7678 \\\hline
    \end{tabular}
    \caption{Baseline model performances on original validation set}
    \label{tab:my_label}
\end{table}

\subsubsection{BERTweet Embedding Extraction}
As mentioned above, we experiment different ways to extract embeddings after feeding Tweets into BERTweet model. Table 3 shows the results of these implementations on original validation set.

\begin{table}[ht]
    \centering
    \begin{tabular}{|c|c|}
        \hline
        \bf{BERTweet Embedding} & \bf{F1 Score} \\\hline
        Last Layer & 0.8912 \\\hline
        All 12 Layers (concat) & 0.9006 \\\hline
        Last 4 Layers (concat) & 0.8934 \\\hline
        Last 2 Layers (concat) & 0.9001 \\\hline
        Last 2 + First 2 (concat) & 0.9013 \\\hline
        Last + First (concat) & \textbf{0.9045} \\\hline
        Last 2 + Mid 2 (concat) & 0.9012 \\\hline
        Last + Mid (concat) & 0.8836 \\\hline
    \end{tabular}
    \caption{Different BERTweet configurations}
    \label{tab:my_label}
\end{table}

\subsubsection{Global Local BERTweet}
Besides experimenting ways to extract BERTweet embeddings, we also experiment different configurations for our Global Local BERTweet model. Table 4 shows the result of these implementations on original validation set.

\begin{table}[ht]
    \centering
    \fontsize{10}{10}\selectfont
    \begin{tabular}{|c|c|c|c|}
        \hline
        \bf{Global} & \bf{Head} & \bf{Tail} & \bf{F1} \\\hline
        last & last & last & 0.9021 \\\hline
        \makecell{last 4 \\ (concat)} & last & last & 0.9028 \\\hline
        \makecell{last 4 \\ (concat)} & \makecell{last + first\\ (concat)} & first & \textbf{0.9075} \\\hline
        \makecell{last 4 \\ (average)} & \makecell{last + first\\ (concat)} & first & 0.8963 \\\hline
        \makecell{last 2 + \\ first 2 \\ (concat)} & \makecell{last + first 2\\ (concat)} & \makecell{last + first 2\\ (concat)} & 0.9067 \\\hline
    \end{tabular}
    \caption{Different BERTweet configuration}
    \label{tab:my_label}
\end{table}

\subsubsection{Ensembling}
Define $\textbf{p}_i$ (dimension $(1 \times 2)$) to be the predicted softmax vector of model $i$-th for each Tweet, $c$ to be the classes (namely Informative/Uninformative), and $N$ to be the number of models. Let $\textbf{C}$ be a function that takes a softmax vector as an input and returns the corresponding binary classification result as output.\\
The output $o_{mv}$ of majority voting is calculated as follows:
\begin{equation}
    o_{mv} = \underset{c}{\text{argmax}} \sum_{i=1}^{N} \textbf{C}(p_i)  
\end{equation}\\
The output $o_{a}$ of averaging is calculated as follows:
\begin{equation}
    o_{a} = \underset{c}{\text{argmax}} \frac{1}{N} \sum_{i=1}^{N} p_i   
\end{equation}

We ensemble all the models shown in Table 3 and Table 4 by doing majority voting and averaging softmax vectors. The results on original validation set are summarized in Table 5.

\begin{table}[ht]
    \centering
    \begin{tabular}{|c|c|}
        \hline
        \bf{Ensembling Method} & \bf{F1} \\\hline
        Majority Voting & \textbf{0.9130} \\\hline
        Averaging & 0.9111 \\\hline
    \end{tabular}
    \caption{Ensembling performance}
    \label{tab:my_label}
\end{table}
\subsubsection{Final Evaluation}
During final evaluation phrase, we used the Majority votted prediction of our BERTweet models after training on the re-splitted training set and got the F1 Score of 0.8991 on the hidden test set, which ranked 12 over 56 participated teams. The first team got the corresponding score of 0.9096.

\subsection{Additional Works}
To investigate our assumption that Tweet length does affect classification result, we analyze the Tweets in the given dataset and come up with an idea to choose the best models for ensembling while dealing with Tweets within a particular length. In particular, we divide the Tweets sequence into 3 categories: short Tweets ($0 - 22$ words), medium Tweets ($23 - 44$ words), long Tweets ($> 44$ words). For each category, we choose 7 models that have the most correct predictions on our training set and use these models for predictions. With this, we gain 0.9182 F1-Score on the original validation set. Indeed, the reported result shows that the selective ensembling of BERTweet models based tailor-trained for a certain range of input Tweet length does boost classification performance. 

\section{Conclusion}

In this paper, we proposed a system that carries out the automatic identification of informative versus uninformative tweets. While this system is simple, it has leveraged recent advances and state-of-the-art results in natural language processing and deep learning, namely BERT-based models. For our future work, we will augment this system so that it can work for various forms of information circulating on social media such as Facebook status, Reddit post, Instagram caption, etc.

\bibliographystyle{acl_natbib}
\bibliography{main}

\end{document}